\documentclass[conference]{IEEEtran}
\IEEEoverridecommandlockouts
\usepackage{url}
\usepackage{cite}
\usepackage{amsmath,amssymb,amsfonts}
\usepackage{algorithmic}
\usepackage{graphicx}
\usepackage{textcomp}
\usepackage{xcolor}
\usepackage{booktabs}
\usepackage{soul}
\usepackage{xcolor}
\usepackage{multirow}

\def\BibTeX{{\rm B\kern-.05em{\sc i\kern-.025em b}\kern-.08em
    T\kern-.1667em\lower.7ex\hbox{E}\kern-.125emX}}
\begin{document}

\title{AccentCL: Robust Accent Classification with Incremental Expansion\
\thanks{This work was supported by the Intelligence Advanced Research Projects Activity (IARPA) via Department of Interior/Interior Business Center (DOI/IBC) contract number 140D0424C0066. The U.S. Government is authorized to reproduce and distribute reprints for Governmental purposes notwithstanding any copyright annotation thereon. Disclaimer: The views and conclusions contained herein are those of the authors and should not be interpreted as necessarily representing the official policies or endorsements, either expressed or implied, of IARPA, DOI/IBC, or the U.S. Government.
Project page: \texttt{https://psi-tamu.github.io/AccentCL/}}
}

\author{
\IEEEauthorblockN{Mu-Ruei Tseng, Waris Quamer, Ghady Nasrallah \& Ricardo Gutierrez-Osuna}
\IEEEauthorblockA{
\textit{Department of Computer Science and Engineering} \\
\textit{Texas A\&M University}\\
College Station, TX 77840, USA \\
\{mtseng, quamer.waris, ghadynasrallah, rgutier\}@tamu.edu \\
}
}

\maketitle

\begin{abstract} 
Accent classifiers are typically trained with a fixed label inventory and cannot accommodate new accent categories as new data becomes available. Moreover, accented speech corpora often exhibit substantial class imbalance and/or domain shift due to differences in recording conditions across corpora. We present AccentCL, a class-incremental learning framework for English accent classification that is robust to class imbalance and cross-corpus domain shift. AccentCL extracts multi-layer representations from a frozen Whisper-Large-v3 encoder, optimized with an imbalance-aware cross-entropy loss to reduce bias toward the majority accent classes and a domain mean alignment loss that minimizes distributional mean shift across training corpora. The label space is then expanded via replay-based continual learning, using the frozen base model for knowledge retention and an old-to-new margin loss to reduce overprediction on newly added classes. On a five-class accent classification task, AccentCL achieves 77.1\% balanced accuracy and a 76.9\% macro-averaged F1 score. We further evaluate the model's ability to incrementally incorporate two new accent categories: Spanish-accented and Chinese-accented English. When adding Spanish-accented English to the pretrained model, AccentCL attains an F1 of 83.3\% on the new class while retaining 77.3\% balanced accuracy on the base classes. When subsequently adding Chinese-accented English, it achieves 61.8\% F1 on the new class while preserving 77.6\% balanced accuracy on the previously learned classes. These results show that AccentCL enables robust regional accent classification while allowing new accent categories to be added without full retraining.
\end{abstract}


\begin{figure*}[t]
    \centering
    \includegraphics[width=0.85\linewidth]{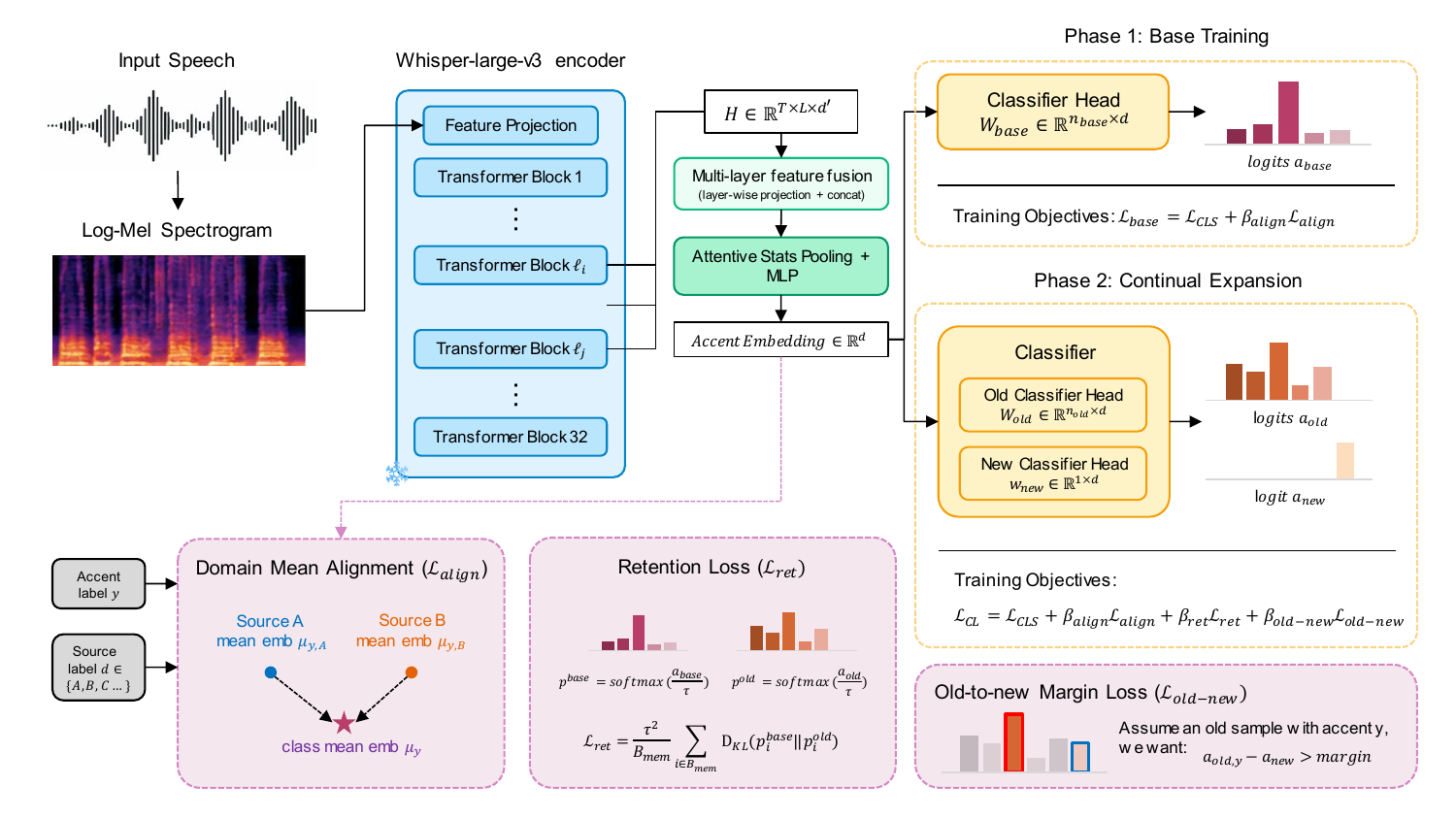}
    \vspace{-15pt}
    \caption{Model architecture of AccentCL. A frozen Whisper-Large-v3 encoder produces multi-layer features that are fused into an accent embedding. Phase~1 trains the regional classifier with domain mean alignment, and Phase~2 expands it with replay, retention, and an old-to-new margin loss.}
    \label{fig:thumbnail}
    \vspace{-15pt}
\end{figure*}

\section{Introduction}
The ability to identify regional or non-native varieties of speech  enables a range of applications, including speaker population analysis, evaluation of accented speech synthesis/conversion systems~\cite{vevo, quamer2026phonos}, and measurement of performance disparities in downstream speech models~\cite{prinos2024speaking}. Further, accent-related variation carries acoustic and phonetic cues (e.g., in pronunciation, rhythm, and prosody) that can affect the performance of automatic speech recognition, voice conversion, and spoken language understanding systems. As speech technologies are increasingly deployed across diverse speaker populations, developing accent classifiers that are both robust and extensible has become an important research direction.

Recent accent classifiers~\cite{zuluaga2023commonaccent, feng2026voxlect} benefit from large pretrained speech encoders and public speech corpora, but often rely on strong assumptions about the underlying data. First, accent labels are typically defined along political or national boundaries (e.g., American vs. Canadian English) rather than geographic or cultural lines, even though intra-national accent variation can be substantial — differences between Southern and Northeastern US accents, for instance, are often more pronounced than those between American and Canadian English. Second, accent label distributions tend to be long-tailed, with some categories represented by substantially fewer samples than others. Third, training samples are frequently pooled across multiple corpora that differ in recording conditions, speaker demographics, and collection protocols, introducing corpus-level domain shift. Finally, these models do not allow new accent categories to be added after a classifier has already been deployed. Retraining from scratch to accommodate new classes is computationally costly and often impractical when the original training data are no longer available or when only limited replay of old-class samples can be retained.

We propose \textsc{AccentCL}, a framework for robust, class-incremental English accent classification. AccentCL first trains an accent classifier by merging closely related fine-grained accent labels into broader regional groups that reflect geographic influences, reducing ambiguity arising from acoustically similar or inconsistently annotated labels. The classifier operates on multi-layer representations extracted from a frozen Whisper-Large-v3 encoder and is trained with class-prior adjustment and corpus-mean alignment to improve robustness to class imbalance and cross-corpus domain shift, respectively. AccentCL then expands the label space by incorporating new accent classes, using a small replay memory of previously learned accents stratified across corpora. During continual training, the updated model is regularized against the frozen base classifier for knowledge retention and uses an old-to-new margin loss to reduce overprediction of newly added classes on samples from previously learned accents. Together, these components allow the classifier to learn new accent categories while preserving performance on previously learned ones.

Our main contributions are:
\begin{itemize}
\item We formulate English accent classification as a robust class-incremental learning problem, one that must contend with ambiguous accent boundaries, class imbalance, domain heterogeneity, and limited replay.
\item We show that multi-layer representations from a frozen speech foundation model can be adapted for robust regional accent classification through imbalance-aware training and class-conditional domain mean alignment.
\item We introduce a replay-based continual training objective that combines domain-stratified memory, frozen-model retention, and an old-to-new margin loss, enabling new accent classes to be added while preserving performance on old ones.
\end{itemize}

\section{Related work}
\subsection{Accent classification with speech foundation models}

Accent classification aims to identify a speaker's regional or non-native accent from speech. Recent systems commonly build on pretrained speech representations, including wav2vec~2.0~\cite{baevski2020wav2vec}, XLS-R~\cite{babu2021xls}, WavLM~\cite{chen2022wavlm}, and Whisper~\cite{radford2022whisper}. These models yield contextual frame-level representations that capture acoustic, phonetic, and speaker-related information~\cite{yang23v_interspeech}, which can then be aggregated via utterance-level pooling and passed to a classification head.

CommonAccent~\cite{zuluaga2023commonaccent} is a pretrained English accent classifier trained on CommonVoice~\cite{ardila2020common} with a fixed 16-class accent inventory, but its single-domain training data and fine-grained label space may limit how well it generalizes to other domains. In our evaluation, many low-resource and out-of-domain accents are confused with related labels or absorbed into dominant classes, which motivates our use of a regional label mapping together with domain-aware training. Voxlect~\cite{feng2026voxlect} introduces a larger mixed-corpus benchmark for English dialect and regional language classification, evaluating several speech foundation models with LoRA fine-tuning; its strongest configuration, based on Whisper-Large-v3, serves as a strong reference model for accent-related classification. However, Voxlect remains a fixed-label classifier and is not built to accommodate new accent categories after training.

AccentCL addresses this limitation by first training a robust regional accent classifier and then extending it through class-incremental learning. Our model also draws on the layer-wise structure of speech foundation models: prior work shows that different encoder layers capture different types of information, including acoustic, phonetic, speaker, and linguistic properties~\cite{chen2022wavlm}. Since accent cues span pronunciation, phoneme realization, rhythm, and prosody, AccentCL aggregates representations across multiple Whisper encoder layers rather than relying on the final layer alone.

\subsection{Class-Incremental Learning}
Class-incremental learning addresses the problem of extending a model with new classes while preserving performance on previously learned ones. Unlike task-incremental learning~\cite{van2019three}, the model must predict over a unified label space at test time, which makes this setting especially prone to catastrophic forgetting and bias toward recently added classes.

Existing methods typically reduce forgetting through replay, which mixes stored old-class examples with new-class data, or through regularization and distillation, which encourage the updated model to stay close to the previous model~\cite{kirkpatrick2017overcoming,li2017learning, rebuffi2017icarl,li2024retention}. This line of work is directly relevant to accent classification, since new accent categories may need to be introduced after a classifier has already been trained. However, fine-tuning directly on the new accent can bias the classifier toward that class and degrade previously learned accents. 

AccentCL supports incremental accent-class expansion by combining new-accent data with memory of previous accents drawn from multiple corpora. During training, it preserves the earlier model's behavior on old-accent examples and suppresses the tendency to misclassify old accents as the newly added class, limiting degradation on earlier learned accents.


\section{Problem formulation}

We study accent classification as a robust class-incremental learning problem. Given an accent classifier trained on a set of ``old" accent classes $\mathcal{Y}_{\mathrm{old}}$, the goal is to add a ``new" accent class $y_{\mathrm{new}}$ while preserving performance on the old classes. 
Before each incremental step, we construct a replay memory $\mathcal{D}_{\mathrm{mem}}$ from the previously learned classes. During
incremental training, the updated classifier $f_{\theta}$ is trained on
$
\mathcal{D}_{\mathrm{train}}
=
\mathcal{D}_{\mathrm{new}}
\cup
\mathcal{D}_{\mathrm{mem}}
$, where $\mathcal{D}_{\mathrm{new}}$ contains training utterances from the newly added accent class. Each training example is represented as $(x_i,y_i,d_i)$, where $x_i$ is an utterance, $y_i\in\mathcal{Y}$ is the accent label, and $d_i$ is the domain/corpus label. 
The objective is to learn a function $f_{\theta}$ that recognizes the new accent class, retains performance on old classes, and remains robust across heterogeneous speech corpora.

\section{Method}
\subsection{Phase 1: Base regional accent classifier}
\label{sec:base_regional_classifier}

The base classifier provides recognition capabilities on the ``old" accents before expanding to new classes. We define an accent label space $\mathcal{Y}_{\mathrm{base}}$ by collapsing the 16 fine-grained accent categories in CommonAccent~\cite{zuluaga2023commonaccent} into 5 broad accent categories: North American (American and Canadian), British Isles (British, Irish, Scottish, and Welsh), Australasian (Australian and New Zealand), South Asian (Indian), and Southeast Asian (Malaysian and Singaporean). This grouping preserves broad regional structure while avoiding unreliable distinctions between fine-grained labels with limited training data.

\paragraph{Multi-layer accent encoder}
Given an input waveform $x$, we use the encoder of a pretrained Whisper-Large-v3 model~\cite{radford2022whisper} as the acoustic feature extractor. We train only the downstream projection, pooling, and classification modules. Motivated by prior findings that intermediate layers of speech foundation models retain useful acoustic and phonetic information~\cite{chen2022wavlm}, we extract hidden states from multiple Whisper encoder layers rather than using only the final layer.

Namely, we select encoder layers $\mathcal{S}=\{16,20,24,28\}$, an evenly spaced middle-to-late subset chosen to cover multiple representation depths while limiting computation. For each layer $\ell \in \mathcal{S}$, let $H^{(\ell)}=\{h_t^{(\ell)}\}_{t=1}^{T}$ denote the corresponding frame-level hidden sequence. We apply a layer-specific projection head $q_{\ell}(\cdot)$ to reduce the encoder's hidden dimension and map all selected layers to the same dimensionality:
\vspace{-5pt}
\begin{equation}
    \tilde{h}_t^{(\ell)} = q_{\ell}\!\left(h_t^{(\ell)}\right), \qquad \tilde{h}_t^{(\ell)} \in \mathbb{R}^{P},
\vspace{-5pt}
\end{equation}
where $P=256$ in our implementation. Each projection head consists of layer normalization followed by a linear projection, a ReLU activation, and dropout.

The projected layer representations are then concatenated at each frame to form a multi-level accent representation, i.e. $
    u_t = \mathrm{Concat}_{\ell \in \mathcal{S}} \left(\tilde{h}_t^{(\ell)}\right)$, where $u_t \in \mathbb{R}^{|\mathcal{S}|P}$.
The resulting sequence $\{u_t\}_{t=1}^{T}$ is passed to the attentive statistics pooling~\cite{okabe2018attentive}, which computes attention-weighted mean and standard-deviation statistics over time. The pooled representation is fed into an MLP to obtain a compact accent embedding, which is then passed to a linear classifier to predict the regional accent label.

\paragraph{Training objective}
For each training example $(x_i,y_i,d_i)$, the model produces logits $a_i$ and an accent embedding $z_i$. To account for class imbalance, we use logit-adjusted cross-entropy~\cite{menon2020long}. Let $\pi \in \mathbb{R}^{|\mathcal{Y}_{\mathrm{base}}|}$ be the empirical class-prior vector in the training set. The classification loss is: 
\vspace{-5pt}
\begin{equation} 
\mathcal{L}_{\mathrm{cls}} = \frac{1}{\mathcal{B}} \sum_{i\in \mathcal{B}} \ell_{\mathrm{CE}}(a_i + \tau \log \pi, y_i),
\vspace{-5pt}
\end{equation}
where $\tau$ controls the strength of the logit adjustment. Setting $\tau=0$ recovers standard cross-entropy.

We also regularize the learned accent embedding space with a domain mean alignment (DMA) loss. Since the same regional accent may appear in multiple datasets with different recording conditions, speaker pools, or corpus collection protocols, this term encourages embeddings to remain consistent across sources while preserving accent-class separation. For accent class $y$ and domain $d$, let $\mu_{y,d}$ denote the mean embedding of samples, and let $\mu_y$ denote the mean embedding of all samples with label $y$ in the mini-batch. The alignment loss is computed over active class-domain groups:
\vspace{-5pt}
\begin{equation} 
\mathcal{L}_{\mathrm{align}} = \frac{1}{|\mathcal{G}|} \sum_{(y,d)\in\mathcal{G}} \left\| \mu_{y,d} - \mu_y \right\|_2^2 , 
\vspace{-5pt}
\end{equation} 
where $\mathcal{G}$ is the set of active class-domain groups in the mini-batch, defined as groups with at least $m$ samples whose accent class is represented in at least two domains. We set $m=2$ in our experiments.  The final base-training objective is $\mathcal{L}_{\mathrm{base}} = \mathcal{L}_{\mathrm{cls}} + \beta_{\mathrm{align}}\mathcal{L}_{\mathrm{align}}$,
where $\beta_{\mathrm{align}}$ controls the strength of the alignment term.



\subsection{Phase 2: Continual expansion}

To add a new accent class, we expand the classifier head of the base model $f_{\mathrm{base}}$ from $|\mathcal{Y}_{\mathrm{old}}|$ outputs to
$|\mathcal{Y}_{\mathrm{old}}|+1$ outputs. The classifier's weights for the old classes are copied from $f_{\mathrm{base}}$, while the new-class weights are randomly initialized. The expanded model, denoted by $f_{\theta}$, is then trained on the union of the new-class data and the replay memory, i.e.
$
\mathcal{D}_{\mathrm{train}}
=
\mathcal{D}_{\mathrm{new}}
\cup
\mathcal{D}_{\mathrm{mem}}
$.

\paragraph{Domain-stratified replay memory}
For class-incremental training, we construct a fixed replay memory from the old accent classes. Let $\mathcal{Y}_{\mathrm{old}}$ denote the old label space. For each old class $k\in\mathcal{Y}_{\mathrm{old}}$, we select
$
    K_k = \min\left(K, |\mathcal{D}^{\mathrm{old}}_{k}|\right)
$
utterances, where $K$ is the maximum number of replay samples per class and $\mathcal{D}^{\mathrm{old}}_{k}$ denotes the old-training utterances with label $k$. Within each class, we stratify sampling by domain/corpus. We divide the $K_k$ selections as evenly as possible across the domains represented in $\mathcal{D}^{\mathrm{old}}_{k}$ and sample without replacement from each domain. If a domain has too few utterances, the remaining samples are drawn at random from the unused utterances of the same class. The resulting class-specific memories are combined to form
$
\mathcal{D}_{\mathrm{mem}}
=
\bigcup_{k\in\mathcal{Y}_{\mathrm{old}}}
\mathcal{D}_{\mathrm{mem}}^{k}
$.
This yields a replay memory that is balanced across old accent classes while maintaining source diversity within each class.

\paragraph{Retention-aware continual training}
Let $\mathcal{Y}=\mathcal{Y}_{\mathrm{old}}\cup\{y_{\mathrm{new}}\}$ denote the expanded label space, where $y_{\mathrm{new}}$ is the newly added accent class. Given utterance $x_i$, the expanded model produces logits
$
a_i = f_{\theta}(x_i) \in \mathbb{R}^{|\mathcal{Y}|}.
$
The primary objective is cross-entropy $\mathcal{L}_{\mathrm{CE}}$ over the expanded label space.

To reduce forgetting on old accents, we keep a frozen copy of the base model $f_{\mathrm{base}}$ and use it as a teacher on replay memory samples. Let $\mathcal{B}_{\mathrm{mem}}\subseteq\mathcal{B}$ denote the old-class replay samples in the mini-batch, and let $C_{\mathrm{old}}=|\mathcal{Y}_{\mathrm{old}}|$. Since the expanded classifier contains an additional new-class output, we restrict its logits to the old label space when computing the retention loss. For replay sample $i$, let
$
a_i^{\mathrm{old}} = a_i[\mathcal{Y}_{\mathrm{old}}]
$ and
$
a^{\mathrm{base}}_i
=
f_{\mathrm{base}}(x_i)
\in
\mathbb{R}^{C_{\mathrm{old}}}
$
denote the old-class logits of the expanded model and  the logits produced by the frozen base model, respectively. We use temperature-scaled knowledge distillation~\cite{hinton2015distilling} to preserve the base model's old-class prediction distribution:
\begin{equation}
p^{\mathrm{base}}_i
=
\mathrm{softmax}
\left(
\frac{a^{\mathrm{base}}_i}{\tau}
\right),
\qquad
p^{\mathrm{old}}_i
=
\mathrm{softmax}
\left(
\frac{a^{\mathrm{old}}_i}{\tau}
\right),
\end{equation}
where $\tau$ is the distillation temperature. The retention loss is defined as:
\begin{equation}
\mathcal{L}_{\mathrm{ret}}
=
\tau^2
\frac{1}{|\mathcal{B}_{\mathrm{mem}}|}
\sum_{i \in \mathcal{B}_{\mathrm{mem}}}
D_{\mathrm{KL}}
\left(
p^{\mathrm{base}}_i
\;\|\;
p^{\mathrm{old}}_i
\right).
\end{equation}
This loss encourages the expanded model to preserve the frozen base model's
old-accent decision structure on replay samples, while the supervised
cross-entropy loss allows the model to learn the newly added accent.

Because $\mathcal{L}_{\mathrm{ret}}$ ignores the newly added class logit, an old replay sample can still receive a high new-class score even when its old-class logits are preserved. Complementary to retention, we add an old-to-new margin loss that directly suppresses new-class overprediction on replay memory samples. For each replay sample, the loss encourages the true old-class logit to exceed the new-class logit by a margin $m$:
\vspace{-5pt}
\begin{equation}
\mathcal{L}_{\mathrm{old,new}}
=
\frac{1}{|\mathcal{B}_{\mathrm{mem}}|}
\sum_{i\in\mathcal{B}_{\mathrm{mem}}}
\max
\left(
0,\,
a_{i,y_{\mathrm{new}}}
-
a_{i,y_i}
+
m
\right).
\vspace{-5pt}
\end{equation}
The final continual-training objective is: 
$\mathcal{L}_{\mathrm{CL}}
=
\mathcal{L}_{\mathrm{CE}}
+
\beta_{\mathrm{align}}\mathcal{L}_{\mathrm{align}} 
+
\beta_{\mathrm{ret}}\mathcal{L}_{\mathrm{ret}}
+
\beta_{\mathrm{old,new}}\mathcal{L}_{\mathrm{old,new}}$ .
Here, $\beta_{\mathrm{align}}$, $\beta_{\mathrm{ret}}$, and
$\beta_{\mathrm{old,new}}$ control the strengths of domain alignment, retention, and old-to-new margin regularization, respectively.

\begin{table*}[ht!]
\centering
\setlength\tabcolsep{2pt}
\caption{
Regional accent classification results on the shared 5-class regional label space. Predictions from CommonAccent and Voxlect are mapped to the same regional label space for evaluation. For models with a larger output label space, predictions outside the evaluated 5-class label set are retained and counted as errors. All metrics are reported in percentages.
}
\label{tab:phase1_regional}
\resizebox{0.85\linewidth}{!}{
\begin{tabular}{lcccccc}
\toprule
\textbf{Method}
& \textbf{All Acc} ($\uparrow$)
& \textbf{All Bal Acc} ($\uparrow$)
& \textbf{All Macro-F1} ($\uparrow$)
& \textbf{OOD Acc} ($\uparrow$)
& \textbf{OOD Bal Acc} ($\uparrow$)
& \textbf{OOD Macro-F1} ($\uparrow$) \\
\midrule
CommonAccent~\cite{zuluaga2023commonaccent}
& 56.0 & 48.3 & 48.6 & 79.0 & 53.6 & 58.2 \\
Voxlect~\cite{feng2026voxlect}
& 64.8 & 65.8 & 68.1 & 83.8 & 77.5 & 82.6 \\
\midrule
\textbf{AccentCL base}
& \textbf{76.0} & 77.1 & \textbf{76.9} & \textbf{89.7} & \textbf{79.6} & \textbf{83.0} \\
\quad w/o logit adjustment
& 75.8 & 76.0 & 76.9 & 87.6 & 77.6 & 80.1 \\
\quad w/o DMA
& 75.7 & \textbf{77.7} & 75.8 & 84.8 & 72.8 & 75.9 \\
\quad w/o logit adjustment \& DMA
& 75.6 & 76.0 & 76.5 & 81.9 & 65.1 & 69.2 \\
\bottomrule
\end{tabular}
}
\vspace{-10pt}
\end{table*}

\section{Experimental Setup}
\paragraph{Datasets}
\label{sec:datasets}
We construct our accent classification dataset from multiple English speech corpora, including CommonVoice~\cite{ardila2020common},
GLOBE~\cite{wang24b_interspeech}, AESRC~\cite{shi2021accented},
UK-Ireland~\cite{demirsahin-etal-2020-open}, EdAcc~\cite{sanabria2023edinburgh},
L2ARCTIC~\cite{zhao2018l2}, ARCTIC~\cite{kominek2003cmu},
EmiliaYODAS~\cite{he2025emilia}, VoxPopuli~\cite{wang2021voxpopuli}, and
EMIME~\cite{wester2010emime}. We assign each utterance to a regional accent group following our collapsed accent label mapping. The full dataset includes North American (NA), British Isles (BI), Australasian (AUS), South Asian (SA), Southeast Asian (SEA), Spanish (SPA), and Chinese (CHN) accent groups.

\begin{figure}[h]
    \vspace{-10pt}
    \centering
    \includegraphics[width=\linewidth]{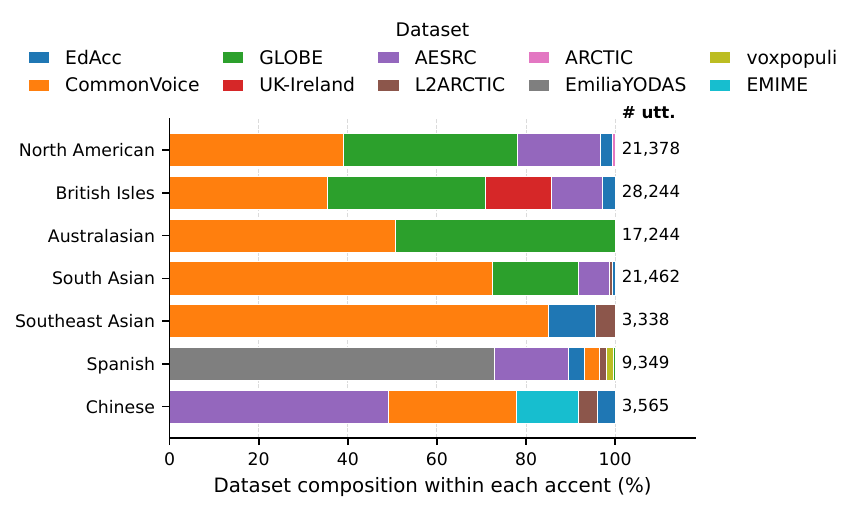}
    \vspace{-15pt}
    \caption{Training dataset composition by accent group. Bars show source-dataset percentages, and right-side numbers indicate utterance counts.}
    \label{fig:accent_dataset_distribution}
    \vspace{-5pt}
\end{figure}

All splits are constructed at the speaker level to prevent speaker overlap among the training, validation, and test sets. Because several corpora contain many utterances from the same speaker, we also limit speaker dominance during training. For each accent group, we first cap each speaker at 50 utterances. We then limit the total number of utterances per accent group to 50,000. To preserve training dataset diversity, we sample across the datasets available for each accent group. This prevents the training subset from being dominated by a small number of frequent speakers or large corpora, while keeping variation in recording conditions across datasets. The resulting training data distribution is shown in Figure~\ref{fig:accent_dataset_distribution}.

\paragraph{Model training} 
We train AccentCL sequentially. All models are trained for 10 epochs with a batch size of 40 using AdamW and cosine learning-rate annealing. All reported results use a single fixed seed of 42. First, we train the base regional accent classifier with a frozen Whisper-Large-v3 encoder and update only the downstream projection layers, attentive statistics pooling module, MLP, and classifier head. The base model uses an initial learning rate of $1\times10^{-3}$, domain mean alignment weight $\beta_{\mathrm{align}}=20$, and logit adjustment temperature $\tau=1.0$. We select the base checkpoint using validation macro-F1. 

We then perform continual training by adding Spanish-accented English as a new class. The classifier head is expanded by copying the old-class weights and initializing a new output unit. The expanded model is trained with new-class data and source-stratified replay memory from the old classes. We set the replay budget to $K=2{,}000$ utterances per old accent class, corresponding to approximately 10\% of the original training data. Continual training uses an initial learning rate of $1\times10^{-4}$, $\beta_{\mathrm{align}}=20$,
retention temperature $\tau=2$, and retention weight $\beta_{\mathrm{ret}}=1.0$.
The margin loss uses $\beta_{\mathrm{old,new}}=0.2$ and margin $m=0.3$.

We repeat the same procedure to add Chinese-accented English on top of the base-plus-Spanish model. For continual training stages, checkpoint selection uses the harmonic mean between old-class macro-F1 and new-class F1: 

\begin{equation}
H_{\mathrm{old,new}} = \frac{ 2 F^{\mathrm{old}}_{\mathrm{macro}} F^{\mathrm{new}} }{ F^{\mathrm{old}}_{\mathrm{macro}} + F^{\mathrm{new}} + \epsilon },
\end{equation} 
which balances old-class retention and new-class learning, instead of selecting models that perform well on only one side of the incremental task.

\paragraph{Baselines}
\label{sec:baselines}
We compare AccentCL against two existing accent classification systems:  \textbf{CommonAccent}~\cite{zuluaga2023commonaccent} is a wav2vec2-XLSR-based English accent classifier trained on CommonVoice. \textbf{Voxlect}~\cite{feng2026voxlect} evaluates several architectures for English dialect classification and reports the best performance with Whisper-Large-v3. We therefore use its Whisper-Large-v3 variant as the Voxlect baseline.  Since CommonAccent, Voxlect, and AccentCL use different output label spaces, all models are evaluated under the shared regional accent label mapping described in Section~\ref{sec:datasets}. Predictions outside the evaluated label set are retained and counted as errors.  For class-incremental experiments, we also report \textbf{Frozen Base} model before adding each new class. This gives the original performance on the old classes and is used as reference for measuring forgetting. We also include \textbf{joint retraining} on all available data as an upper bound and \textbf{naive new-class-only fine-tuning} with cross-entropy and no replay as a lower bound.
\paragraph{Evaluation metrics} \label{sec:evaluation_metrics} We report accuracy, balanced accuracy, and macro-F1 for regional accent classification. Balanced accuracy is computed as the mean per-class recall. 
Macro-F1 is the unweighted average of class-wise F1 scores. We report results under two evaluation settings. \textbf{All} evaluates all test utterances from our constructed benchmark whose labels fall within the target regional label set. \textbf{OOD} evaluates source-held-out generalization on Speech Accent Archive~\cite{weinberger2011speech} and IDEA~\cite{meier1998idea}, neither of which is used as a training source for any model.

For class-incremental learning, we report new-class F1 and old-class balanced accuracy, where old classes are those learned before the current incremental step. To measure forgetting, we compute the per-class accuracy change between the expanded model and the model before adding the new accent, i.e.
$
\delta_k =
\mathrm{Acc}^{\mathrm{after}}_k -
\mathrm{Acc}^{\mathrm{before}}_k .
$
We then report the average forgetting over all old classes and the worst-class
change:
\vspace{-5pt}
\begin{equation}
\Delta_{\mathrm{avg}}
=
\frac{1}{|\mathcal{Y}_{\mathrm{old}}|}
\sum_{k\in\mathcal{Y}_{\mathrm{old}}}
\min(0,\delta_k),
\qquad
\Delta_{\mathrm{worst}}
=
\min_{k\in\mathcal{Y}_{\mathrm{old}}}
\delta_k .
\vspace{-5pt}
\end{equation}
More negative values indicate stronger forgetting.

\section{Results}
\label{sec:results}

\subsection{Robust regional accent classification}

Table~\ref{tab:phase1_regional} reports the performance of the 5-class base regional accent classifier. We evaluate each model on the full test set and on a source-held-out subset to assess generalization across recording sources. On the full evaluation set, AccentCL base achieves an accuracy of \textbf{76.0\%}, balanced accuracy of \textbf{77.1\%}, and macro-F1 of \textbf{76.9\%}. Compared with the mapped Voxlect baseline, AccentCL base improves accuracy by \textbf{+11.2} points, balanced accuracy by \textbf{+11.3} points, and macro-F1 by \textbf{+8.8} points. These results show that the proposed regional classifier provides stronger overall and class-balanced performance under the shared five-class accent label space. 

On the source-held-out subset, AccentCL base achieves an OOD accuracy of \textbf{89.7\%}, balanced accuracy of \textbf{79.6\%}, and macro-F1 of \textbf{83.0\%}. Compared with Voxlect, AccentCL improves OOD accuracy by \textbf{+5.9} points, balanced accuracy by \textbf{+2.1} points, and macro-F1 by \textbf{+0.4} points. Although the OOD macro-F1 gain is modest, AccentCL maintains strong class-balanced performance while improving overall accuracy on held-out recording sources. Ablation results in Table I further show that both logit adjustment and DMA improve OOD robustness, with DMA having the larger individual effect and their combination providing the strongest generalization to unseen recording sources.

\begin{table}[h!]
\centering
\setlength\tabcolsep{2pt}
\caption{
Class-incremental accent learning results and ablations. Old BAcc. is computed
over the classes learned before each incremental step. $\Delta_{\mathrm{avg}}$
and $\Delta_{\mathrm{worst}}$ denote old-class accuracy changes relative to the old model.
}
\label{tab:phase2_cl}
\resizebox{0.85\linewidth}{!}{
\begin{tabular}{lcccc}
\toprule
\textbf{Method}
& \textbf{New F1} $\uparrow$
& \textbf{Old BAcc} $\uparrow$
& $\boldsymbol{\Delta_{\mathrm{avg}}}$ $\uparrow$
& $\boldsymbol{\Delta_{\mathrm{worst}}}$ $\uparrow$ \\
\midrule
\multicolumn{5}{l}{\textit{Base model: 5 regional classes}} \\
Frozen Base & -- & 77.1 & -- & -- \\
\midrule
\multicolumn{5}{l}{\textit{Step 1: $5 \rightarrow 6$, adding Spanish-accented English}} \\
Voxlect & 52.7 & 66.0 & -- & --  \\
Replay only & 81.4 & 76.2
         & -1.6 & -4.4 (BI) \\
Replay + retention & 82.8 & 77.1
         & \textbf{-0.7} & \textbf{-3.6 (BI)} \\
Replay + margin & 82.2 & 76.1
         & -1.8 & -4.6 (BI) \\
AccentCL & \textbf{83.3} & \textbf{77.3}
         & -0.8 & -3.9 (BI) \\
\midrule
\multicolumn{5}{l}{\textit{Step 2: $6 \rightarrow 7$, adding Chinese-accented English}} \\
Voxlect & -- & 63.7 & -- & -- \\
Replay only & 41.9 & 75.1
         & -3.7 & -7.5 (SPA) \\
Replay + retention & 48.5 & 75.9
         & -2.9 & -5.7 (SPA) \\
Replay + margin & 59.9 & 77.1
         & \textbf{-1.4} & \textbf{-3.8 (NA)} \\
AccentCL & \textbf{61.8} & \textbf{77.6}
         & -1.7 & -4.2 (NA) \\
\bottomrule
\end{tabular}
}
\vspace{-15pt}
\end{table}

\subsection{Class-incremental accent expansion}

We next evaluate whether the regional accent inventory can be expanded without retraining the full model from scratch. Starting from the five-class base model, we incrementally add two accent classes:
first \textit{Spanish-accented English}, and then \textit{Chinese-accented English}, yielding a class-incremental sequence of $5 \rightarrow 6 \rightarrow 7$ classes.

\paragraph{Incremental performance}
Table~\ref{tab:phase2_cl} reports the class-incremental results. Voxlect is included as a fixed-label reference model rather than a continual-learning baseline. After adding Spanish-accented English, AccentCL achieves the best new-class F1 (\textbf{83.3}) and old-class balanced accuracy (\textbf{77.3\%}), while showing only a small average old-class drop (\textbf{-0.8} points). The retention-only variant gives the smallest forgetting in this step, suggesting that matching the previous classifier's predictions helps preserve the performance of the original five regional classes. After adding Chinese-accented English, AccentCL again obtains the best new-class F1 (\textbf{61.8}) and old-class balanced accuracy (\textbf{77.6\%}). The margin-only variant has the smallest degradation on old classes in this step, indicating that reducing confusions between old accents and the new class is important when adding Chinese-accented English. Overall, the ablations show that retention and the margin loss contribute to the two expansion steps. Combining both gives \textsc{AccentCL} the most consistent balance between learning the new accent and maintaining performance on old accents.

As lower and upper bounds, naive new-class-only fine-tuning and full joint retraining achieve 15.0/0.3 and 89.3/77.7 New F1/Old BAcc, respectively, for the $5\!\rightarrow\!6$ step, and 5.7/5.7 and 70.3/80.0 for the $6\!\rightarrow\!7$ step. These results show that naive fine-tuning suffers severe catastrophic forgetting, while AccentCL approaches full joint retraining using only limited replay.

\begin{figure}[h]
    \centering
    \includegraphics[width=0.9\linewidth]{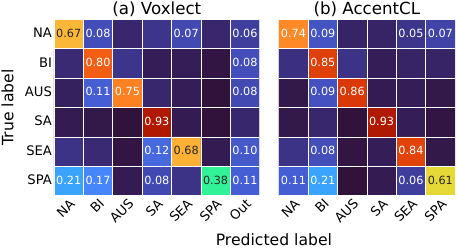}
    \vspace{-10pt}
    \caption{Per-class recall matrices for Voxlect \& AccentCL after adding Spanish-accented English. \textit{Out} denotes predictions outside the evaluated labels.}
    \label{fig:comparison_voxlect_accentcl}
    \vspace{-5pt}
\end{figure}

\paragraph{Qualitative analysis}
Figure~\ref{fig:comparison_voxlect_accentcl} compares per-class recall for Voxlect and AccentCL after adding Spanish-accented English. We evaluate on the held-out speakers from corpora used by both models. AccentCL shows higher recall for the newly added
Spanish-accented English class and for old regional classes, which is consistent with the gains in Table~\ref{tab:phase2_cl}.
    
\begin{figure}[h]
    \centering
    \includegraphics[width=\linewidth]{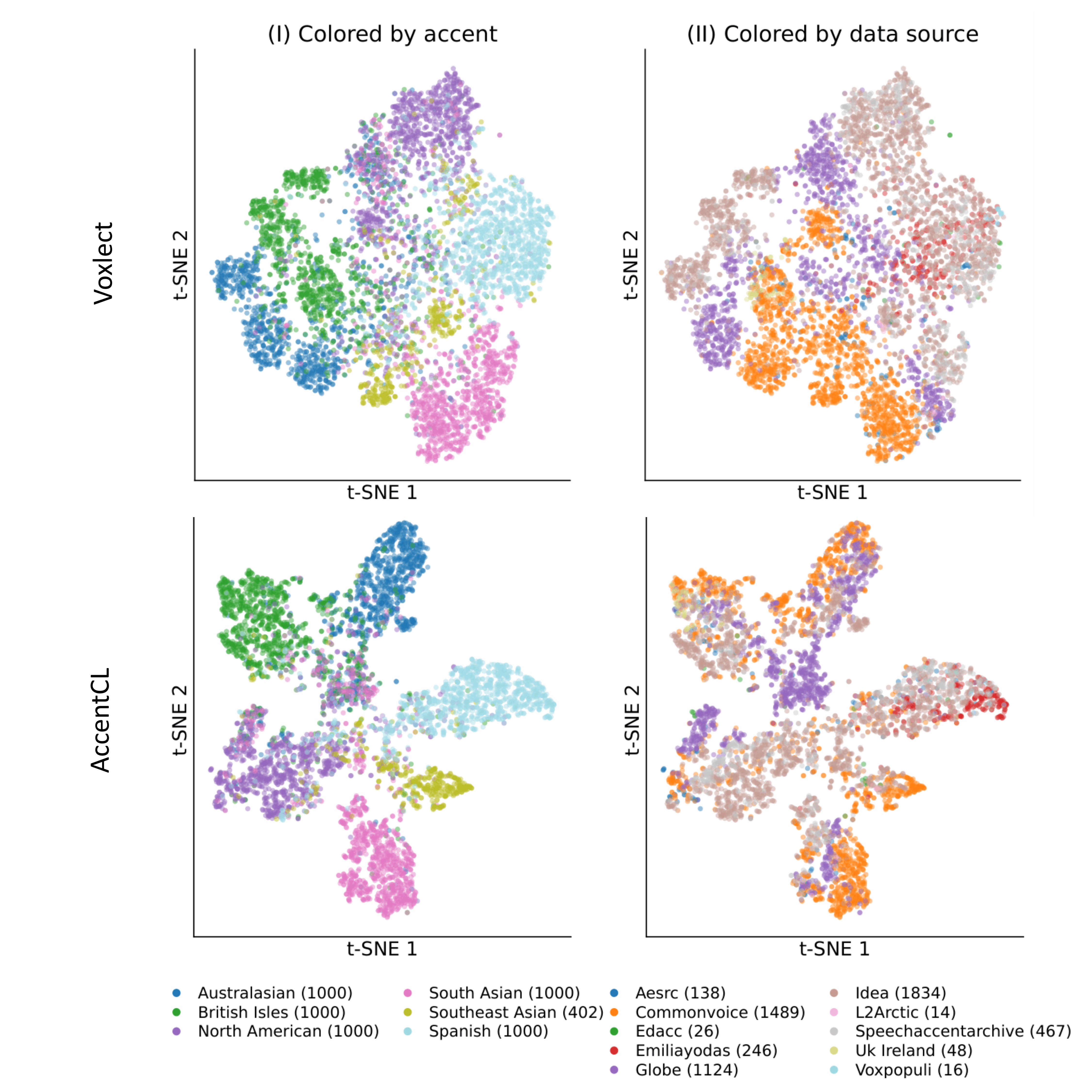}
    \vspace{-15pt}
    \caption{t-SNE of Voxlect (top) and AccentCL embeddings (bottom) after adding Spanish-accented English. Points are colored by accent label (left) and by domain (right). 
    }
    \label{fig:tsne_feat}
    \vspace{-15pt}
\end{figure}

Fig.~\ref{fig:tsne_feat} shows t-SNE visualizations of learned embedding spaces. Compared with Voxlect, AccentCL forms clearer accent groups with weaker domain shifts. For example, Australasian samples in VoxLect are  split into several clusters from different domains, while AccentCL keeps them more compact. Some domain shifts remain, suggesting that AccentCL reduces, but does not fully remove, variation across corpora.

\vspace{-10pt}

\section{Discussion}
We proposed \textsc{AccentCL}, a framework for English accent classification that supports class-incremental expansion. By combining data from new accents with replay memory stratified by dataset, AccentCL adds Spanish- and Chinese-accented English without retraining on the full old dataset. In our setting, only about 10\% of the old training data is retained as replay memory and still maintains balanced accuracy on old accent classes across incremental steps. 

The ablation results show that retention and old-to-new margin regularization provide complementary benefits. Retention preserves the previous classifier's behavior while the margin loss reduces confusion between old accents and the newly added class. New-class F1, balanced accuracy on old classes, and worst-class forgetting further show that AccentCL balances learning new accents with retaining old accents. A limitation of the current study is that continual learning is evaluated over only two expansion steps, using a single class order and random seed. Future work will examine robustness across additional expansion sequences and random seeds.

Compared with the Voxlect state-of-the-art baseline, AccentCL achieves stronger regional accent classification and better adaptation to the expanded label space. These results show that AccentCL provides a practical pipeline for continually expanding accent classifiers to future accent categories without retraining from scratch or substantially degrading previously learned classes.

\section*{Acknowledgments}
Generative AI tools were used only for editing to improve clarity and presentation. All experimental design, implementation, data analysis, and result
interpretation were performed by the authors.


\bibliographystyle{IEEEtran}
\bibliography{ref}

@article{zuluaga2023commonaccent,
  title={CommonAccent: Exploring Large Acoustic Pretrained Models for Accent Classification Based on Common Voice},
  author={Zuluaga-Gomez, Juan and Ahmed, Sara and Visockas, Danielius and Subakan, Cem},
  journal={Interspeech 2023},
  url={https://arxiv.org/abs/2305.18283},
  year={2023}
}

@article{li2024retention,
  title={A Retention-Centric Framework for Continual Learning with Guaranteed Model Developmental Safety},
  author={Li, Gang and Yu, Wendi and Yao, Yao and Tong, Wei and Liang, Yingbin and Lin, Qihang and Yang, Tianbao},
  journal={arXiv preprint arXiv:2410.03955},
  year={2024}
}

@inproceedings{ardila2020common,
  title={Common voice: A massively-multilingual speech corpus},
  author={Ardila, Rosana and Branson, Megan and Davis, Kelly and Kohler, Michael and Meyer, Josh and Henretty, Michael and Morais, Reuben and Saunders, Lindsay and Tyers, Francis and Weber, Gregor},
  booktitle={Proceedings of the twelfth language resources and evaluation conference},
  pages={4218--4222},
  year={2020}
}

@inproceedings{shi2021accented,
  title={The accented english speech recognition challenge 2020: open datasets, tracks, baselines, results and methods},
  author={Shi, Xian and Yu, Fan and Lu, Yizhou and Liang, Yuhao and Feng, Qiangze and Wang, Daliang and Qian, Yanmin and Xie, Lei},
  booktitle={ICASSP 2021-2021 IEEE International Conference on Acoustics, Speech and Signal Processing (ICASSP)},
  pages={6918--6922},
  year={2021},
  organization={IEEE}
}

@article{he2025emilia,
  title={Emilia: A large-scale, extensive, multilingual, and diverse dataset for speech generation},
  author={He, Haorui and Shang, Zengqiang and Wang, Chaoren and Li, Xuyuan and Gu, Yicheng and Hua, Hua and Liu, Liwei and Yang, Chen and Li, Jiaqi and Shi, Peiyang and others},
  journal={IEEE Transactions on Audio, Speech and Language Processing},
  year={2025},
  publisher={IEEE}
}

@inproceedings{feng2026voxlect,
  title={Voxlect: A speech foundation model benchmark for modeling dialects and regional languages around the globe},
  author={Feng, Tiantian and Huang, Kevin and Xu, Anfeng and Shi, Xuan and Lertpetchpun, Thanathai and Lee, Jihwan and Lee, Yoonjeong and Byrd, Dani and Narayanan, Shrikanth},
  booktitle={Proceedings of the 32nd ACM SIGKDD Conference on Knowledge Discovery and Data Mining V. 1},
  pages={2640--2651},
  year={2026}
}

@misc{radford2022whisper,
  doi = {10.48550/ARXIV.2212.04356},
  url = {https://arxiv.org/abs/2212.04356},
  author = {Radford, Alec and Kim, Jong Wook and Xu, Tao and Brockman, Greg and McLeavey, Christine and Sutskever, Ilya},
  title = {Robust Speech Recognition via Large-Scale Weak Supervision},
  publisher = {arXiv},
  year = {2022},
  copyright = {arXiv.org perpetual, non-exclusive license}
}

@inproceedings{okabe2018attentive,
  title     = {{Attentive Statistics Pooling for Deep Speaker Embedding}},
  author    = {Koji Okabe and Takafumi Koshinaka and Koichi Shinoda},
  year      = {2018},
  booktitle = {{Interspeech 2018}},
  pages     = {2252--2256},
  doi       = {10.21437/Interspeech.2018-993},
  issn      = {2958-1796},
}

@article{menon2020long,
  title={Long-tail learning via logit adjustment},
  author={Menon, Aditya Krishna and Jayasumana, Sadeep and Rawat, Ankit Singh and Jain, Himanshu and Veit, Andreas and Kumar, Sanjiv},
  journal={arXiv preprint arXiv:2007.07314},
  year={2020}
}

@article{chen2022wavlm,
  title={Wavlm: Large-scale self-supervised pre-training for full stack speech processing},
  author={Chen, Sanyuan and Wang, Chengyi and Chen, Zhengyang and Wu, Yu and Liu, Shujie and Chen, Zhuo and Li, Jinyu and Kanda, Naoyuki and Yoshioka, Takuya and Xiao, Xiong and others},
  journal={IEEE Journal of Selected Topics in Signal Processing},
  volume={16},
  number={6},
  pages={1505--1518},
  year={2022},
  publisher={IEEE}
}

@inproceedings{wang24b_interspeech,
  title     = {{GLOBE: A High-quality English Corpus with Global Accents for Zero-shot Speaker Adaptive Text-to-Speech}},
  author    = {Wenbin Wang and Yang Song and Sanjay Jha},
  year      = {2024},
  booktitle = {{Interspeech 2024}},
  pages     = {1365--1369},
  doi       = {10.21437/Interspeech.2024-70},
  issn      = {2958-1796},
}

@inproceedings{demirsahin-etal-2020-open,
    title = {{Open-source Multi-speaker Corpora of the English Accents in the British Isles}},
    author = {Demirsahin, Isin and Kjartansson, Oddur and Gutkin, Alexander and Rivera, Clara},
    booktitle = {Proceedings of The 12th Language Resources and Evaluation Conference (LREC)},
    month = may,
    year = {2020},
    pages = {6532--6541},
    address = {Marseille, France},
    publisher = {European Language Resources Association (ELRA)},
    url = {https://www.aclweb.org/anthology/2020.lrec-1.804},
    ISBN = {979-10-95546-34-4},
  }

@inproceedings{sanabria2023edinburgh,
  title={The edinburgh international accents of english corpus: Towards the democratization of english asr},
  author={Sanabria, Ramon and Bogoychev, Nikolay and Markl, Nina and Carmantini, Andrea and Klejch, Ondrej and Bell, Peter},
  booktitle={ICASSP 2023-2023 IEEE International Conference on Acoustics, Speech and Signal Processing (ICASSP)},
  pages={1--5},
  year={2023},
  organization={IEEE}
}

@inproceedings{zhao2018l2,
  title={L2-ARCTIC: A Non-native English Speech Corpus},
  author={Zhao, Guanlong and Sonsaat, Sinem and Silpachai, Alif and Lucic, Ivana and Chukharev-Hudilainen, Evgeny and Levis, John and Gutierrez-Osuna, Ricardo},
  booktitle={Proc. Interspeech 2018},
  pages={2783--2787},
  year={2018}
}

@inproceedings{kominek2003cmu,
  title={The CMU Arctic speech databases},
  author={Kominek, John and Black, Alan W},
  booktitle={Proc. SSW 2004},
  pages={223--224},
  year={2004}
}

@inproceedings{wang2021voxpopuli,
  title={VoxPopuli: A large-scale multilingual speech corpus for representation learning, semi-supervised learning and interpretation},
  author={Wang, Changhan and Riviere, Morgane and Lee, Ann and Wu, Anne and Talnikar, Chaitanya and Haziza, Daniel and Williamson, Mary and Pino, Juan and Dupoux, Emmanuel},
  booktitle={Proceedings of the 59th Annual Meeting of the Association for Computational Linguistics and the 11th International Joint Conference on Natural Language Processing (Volume 1: Long Papers)},
  pages={993--1003},
  year={2021}
}

@techreport{wester2010emime,
  title={The EMIME bilingual database},
  author={Wester, Mirjam},
  year={2010},
  institution={The University of Edinburgh}
}

@article{weinberger2011speech,
  title={The Speech Accent Archive: towards a typology of English accents.},
  author={Weinberger, Steven H and Kunath, Stephen A},
  journal={Language \& Computers},
  volume={73},
  number={1},
  year={2011}
}

@article{meier1998idea,
  title={IDEA: International dialects of English archive},
  author={Meier, Paul and Muller, SM},
  journal={Accessed May},
  volume={17},
  pages={2005},
  year={1998}
}

@article{baevski2020wav2vec,
  title={wav2vec 2.0: A framework for self-supervised learning of speech representations},
  author={Baevski, Alexei and Zhou, Yuhao and Mohamed, Abdelrahman and Auli, Michael},
  journal={Advances in neural information processing systems},
  volume={33},
  pages={12449--12460},
  year={2020}
}

@article{babu2021xls,
  title={XLS-R: Self-supervised cross-lingual speech representation learning at scale},
  author={Babu, Arun and Wang, Changhan and Tjandra, Andros and Lakhotia, Kushal and Xu, Qiantong and Goyal, Naman and Singh, Kritika and Von Platen, Patrick and Saraf, Yatharth and Pino, Juan and others},
  journal={arXiv preprint arXiv:2111.09296},
  year={2021}
}

@article{kirkpatrick2017overcoming,
  title={Overcoming catastrophic forgetting in neural networks},
  author={Kirkpatrick, James and Pascanu, Razvan and Rabinowitz, Neil and Veness, Joel and Desjardins, Guillaume and Rusu, Andrei A and Milan, Kieran and Quan, John and Ramalho, Tiago and Grabska-Barwinska, Agnieszka and others},
  journal={Proceedings of the national academy of sciences},
  volume={114},
  number={13},
  pages={3521--3526},
  year={2017},
  publisher={National Academy of Sciences}
}

@article{li2017learning,
  title={Learning without forgetting},
  author={Li, Zhizhong and Hoiem, Derek},
  journal={IEEE transactions on pattern analysis and machine intelligence},
  volume={40},
  number={12},
  pages={2935--2947},
  year={2017},
  publisher={IEEE}
}

@inproceedings{rebuffi2017icarl,
  title={icarl: Incremental classifier and representation learning},
  author={Rebuffi, Sylvestre-Alvise and Kolesnikov, Alexander and Sperl, Georg and Lampert, Christoph H},
  booktitle={Proceedings of the IEEE conference on Computer Vision and Pattern Recognition},
  pages={2001--2010},
  year={2017}
}

@inproceedings{vevo,
  author       = {Xueyao Zhang and Xiaohui Zhang and Kainan Peng and Zhenyu Tang and Vimal Manohar and Yingru Liu and Jeff Hwang and Dangna Li and Yuhao Wang and Julian Chan and Yuan Huang and Zhizheng Wu and Mingbo Ma},
  title        = {Vevo: Controllable Zero-Shot Voice Imitation with Self-Supervised Disentanglement},
  booktitle    = {{ICLR}},
  publisher    = {OpenReview.net},
  year         = {2025}
}

@article{quamer2026phonos,
  title={PHONOS: PHOnetic Neutralization for Online Streaming Applications},
  author={Quamer, Waris and Tseng, Mu-Ruei and Nasrallah, Ghady and Gutierrez-Osuna, Ricardo},
  journal={arXiv preprint arXiv:2603.27001},
  year={2026}
}

@inproceedings{prinos2024speaking,
  title={Speaking of accent: A content analysis of accent misconceptions in ASR research},
  author={Prinos, Kerri and Patwari, Neal and Power, Cathleen A},
  booktitle={Proceedings of the 2024 ACM Conference on Fairness, Accountability, and Transparency},
  pages={1245--1254},
  year={2024}
}

@article{hinton2015distilling,
  title={Distilling the knowledge in a neural network},
  author={Hinton, Geoffrey and Vinyals, Oriol and Dean, Jeff},
  journal={arXiv preprint arXiv:1503.02531},
  year={2015}
}

@inproceedings{yang23v_interspeech,
  title     = {{What Can an Accent Identifier Learn? Probing Phonetic and Prosodic Information in a Wav2vec2-based Accent Identification Model}},
  author    = {Mu Yang and Ram C. M. C. Shekar and Okim Kang and John H. L. Hansen},
  year      = {2023},
  booktitle = {{Interspeech 2023}},
  pages     = {1923--1927},
  doi       = {10.21437/Interspeech.2023-2254},
  issn      = {2958-1796},
}

@article{van2019three,
  title={Three scenarios for continual learning},
  author={Van de Ven, Gido M and Tolias, Andreas S},
  journal={arXiv preprint arXiv:1904.07734},
  year={2019}
}

\end{document}